\documentclass{article} 
\usepackage{iclr2024_conference, times}
\makeatletter
\def\input@path{{figures/}{sections/}}
\makeatother

\usepackage{amsmath,amsfonts,bm}

\def\eqref#1{equation~\ref{#1}}

\def\1{\bm{1}}

\def\ra{{\textnormal{a}}}

\def\rx{{\textnormal{x}}}

\def\rva{{\mathbf{a}}}

\def\erva{{\textnormal{a}}}

\def\ervx{{\textnormal{x}}}

\def\rmA{{\mathbf{A}}}

\def\vmu{{\bm{\mu}}}
\def\vtheta{{\bm{\theta}}}
\def\va{{\bm{a}}}

\def\ve{{\bm{e}}}

\def\vx{{\bm{x}}}

\def\eva{{a}}

\def\mA{{\bm{A}}}

\def\mH{{\bm{H}}}
\def\mI{{\bm{I}}}
\def\mJ{{\bm{J}}}

\def\mX{{\bm{X}}}

\def\mSigma{{\bm{\Sigma}}}

\DeclareMathAlphabet{\mathsfit}{\encodingdefault}{\sfdefault}{m}{sl}
\SetMathAlphabet{\mathsfit}{bold}{\encodingdefault}{\sfdefault}{bx}{n}
\newcommand{\tens}[1]{\bm{\mathsfit{#1}}}
\def\tA{{\tens{A}}}

\def\tX{{\tens{X}}}

\def\gG{{\mathcal{G}}}

\def\sA{{\mathbb{A}}}
\def\sB{{\mathbb{B}}}

\def\sS{{\mathbb{S}}}

\def\emA{{A}}

\newcommand{\etens}[1]{\mathsfit{#1}}

\def\etA{{\etens{A}}}

\newcommand{\E}{\mathbb{E}}

\newcommand{\R}{\mathbb{R}}

\newcommand{\KL}{D_{\mathrm{KL}}}
\newcommand{\Var}{\mathrm{Var}}

\newcommand{\Cov}{\mathrm{Cov}}

\newcommand{\normltwo}{L^2}
\newcommand{\normlp}{L^p}

\newcommand{\parents}{Pa} 

\usepackage{bbm}
\usepackage{wrapfig}
\usepackage{hyperref}
\usepackage{url}
\usepackage{float}
\usepackage{tabularx}
\usepackage{makecell}
\usepackage{array}
\usepackage{colortbl}
\usepackage{subcaption}
\usepackage{graphicx}
\usepackage{pgfplots}
\pgfplotsset{compat=1.10}
\usepackage{todonotes}
\usepackage{cleveref}
\usepackage{algorithm}
\usepackage{algpseudocode}
\usepackage{verbatim}
\usepackage{multirow}

\usetikzlibrary{positioning, matrix, shapes.geometric, arrows.meta, fit, backgrounds, calc, intersections, fillbetween}
\usepgfplotslibrary{fillbetween}

\algtext*{EndIf}
\algtext*{EndWhile}
\algtext*{EndFor}
\algtext*{EndFunction}
\algtext*{EndProcedure}

\newcolumntype{C}{>{\centering\arraybackslash}X}
\newcolumntype{s}{>{\hsize=.3\hsize\linewidth=\hsize}C}
\newcolumntype{D}{>{\hsize=.4\hsize\linewidth=\hsize}C} 

\title{Offline Learning of Controllable Diverse Behaviors}

\author{
    Mathieu Petitbois\textsuperscript{*,1}, 
    Rémy Portelas\textsuperscript{1},
    Sylvain Lamprier\textsuperscript{2},
    Ludovic Denoyer\textsuperscript{3} \\
    \textsuperscript{1}Ubisoft La Forge
    \textsuperscript{2}University of Angers 
    \textsuperscript{3}H Company\\
}

\iclrfinalcopy 
\begin{document}

\renewcommand{\thefootnote}{\fnsymbol{footnote}}
\footnotetext[1]{Correspondence to \texttt{mathieu.petitbois@ubisoft.com}}

\maketitle

\begin{abstract}
Imitation Learning (IL) techniques aim to replicate human behaviors in specific tasks. While IL has gained prominence due to its effectiveness and efficiency, traditional methods often focus on datasets collected from experts to produce a single efficient policy. Recently, extensions have been proposed to handle datasets of diverse behaviors by mainly focusing on learning transition-level diverse policies or on performing entropy maximization at the trajectory level. While these methods may lead to diverse behaviors, they may not be sufficient to reproduce the actual diversity of demonstrations or to allow controlled trajectory generation. To overcome these drawbacks, we propose a different method based on two key features: a) Temporal Consistency that ensures consistent behaviors across entire episodes and not just at the transition level as well as b) Controllability obtained by constructing a latent space of behaviors that allows users to selectively activate specific behaviors based on their requirements. We compare our approach to state-of-the-art methods over a diverse set of tasks and environments. Project page: \url{https://mathieu-petitbois.github.io/projects/swr/}
\end{abstract}

\section{Introduction}
For several years, Imitation Learning (IL) from diverse pre-generated human demonstrations has found success in learning to solve a diverse set of tasks in sequential decision making scenarios \citep{kumar2022prefer}, showing great promises to improve over traditional methods in many fields such as robotics \citep{mandlekar2021matters}, video-games \citep{shen2020diverse} or even autonomous driving \citep{10.1145/3542945}. For robotics, learning from human experts allows to reach human-level performance without any controller hard coding or expensive interaction with simulated or real environments. Similarly, in video games, it facilitates the training of human-like agents and serves as an alternative to the traditional methods of behavior coding, which are notoriously time-consuming and less effective in producing realistic behaviors. For example, the programming of bots in video games typically relies on scripting techniques such as State Machines \citep{ArtificialIntelligenceforGames,AiGameDevelopment}, Utility Systems \citep{BehavioralMathematicsforGameAI}, Decision Trees \citep{ArtificialIntelligenceAModernApproach,Laird_VanLent_2001}, and Planners \citep{Orkin2006ThreeSA}. These methods not only require considerable time to implement but also struggle to replicate the complex behaviors exhibited by human players, thus failing to enhance the gaming experience or providing realistic bots to test games.

However, learning from diverse human data also presents its challenges. Human data diversity might be driven by a wide range  of objectives, skill levels, hesitations and noisy actions which undermine many traditional imitation learning techniques. While traditional IL techniques focus on training an agent using a dataset of behavioral traces, they traditionally operate under the assumption that the data originates from a single expert policy. This limitation often prevents them from capturing the varied nature of behaviors. Consequently, standard IL methods like Behavioral Cloning (BC) \citep{NIPS1988_812b4ba2} or Generative Adversarial Imitation Learning (GAIL) \citep{ho2016generative} fail to capture the demonstration diversity of multi-modal datasets. In contexts where the goal is to develop a single, efficient policy, this limitation may not be critical. However, in scenarios where capturing diverse behaviors is essential, such as in video games where bots exhibiting varied behaviors enhance the game's realism and engagement, these methods fall short. Therefore, it is crucial to advance imitation learning to the diversity of human behaviors.

The challenge of addressing multi-modality in imitation learning has recently been tackled with different scopes. A first category of methods are considering the capture of diversity at the \textbf{transition-level} \citep{reuss2023goalconditioned, florence2021implicit, pearce2023imitating}: they capture the diversity of actions for each specific state at the current timestep without taking into account long term information. A second category of methods consider diversity at \textbf{trajectory-level} using auto-encoders (AE) \citep{sudhakaran2023skill}, GAIL extensions \citep{hausman2017multimodal, li2017infogail, wang2017robust}  or even diffusion models \citep{janner2022planning, ajay2023is}, conditioning their agents on the whole trajectory. However, transition-level methods suffer from a lack of temporal consistency at the trajectory scale, which might not reproduce the diversity of trajectories present in the demonstrations. Furthermore, while trajectory-level methods capture trajectory-level diversity to some extent, most do so in an online learning framework, allowing interactions with the environments, to perform a Reinforcement Learning (RL) task reward optimization or to maximize the entropy of certain displays of behaviors, without aiming for the reproduction of the actual trajectory distribution. To our knowledge, no methods directly aimed at capturing the real diversity of trajectory in a fully offline setting while ensuring a flexible control ability of the generation process have been designed and evaluated as such.

In response to these challenges, we propose a novel approach designed to capture controllable and diverse behaviors from a dataset of collected traces.  Our main contributions are of the following:
\begin{itemize}
    \item We introduce a model architecture that can capture the human diversity at the trajectory scale through unsupervised learning while being controllable and displaying robustness to stochasticity.
    \item We propose the use of a distance on histograms of generated behaviors to measure diversity reconstruction on a diverse set of human-generated data.
    \item We analyze the performance of our model compared to the baselines on diversity capture, controllabilty and robustness to stochasticity on a diverse set of human generated datasets and environments: A new \textbf{Maze2D} environment as well as modified versions of the datasets provided by the D3IL \citep{sudhakaran2023skill} library.
\end{itemize}

\section{Related Work}
\textbf{Imitation learning and offline reinforcement learning} While the goal of RL is to maximize the cumulated rewards of a given agent in interaction with an environment, the goal of offline RL is to do so by learning from pre-recorded demonstrations with no interaction with the environment. Offline RL often involves the estimation of the quality of actions for off-policy improvement while avoiding value extrapolation due to distributional shift \citep{fujimoto2019offpolicy, kostrikov2021offline, kumar2020conservative}. Imitation learning on the other hand traditionally aims to learn to copy the behavior policy that generated the demonstration dataset, mostly by reproducing its state conditioned action distribution, called behavior cloning \citep{kumar2022prefer} or infering a reward function that the behavior policy should be optimizing, called inverse reinforcement learning (IRL) \citep{ho2016generative}. In our setting, we aim to reproduce not only distribution of actions generated by the behaviors policies but instead the distribution of the trajectories.

\textbf{Multimodal imitation learning at the action-scale} To treat the multimodality of human data, extensions to traditional imitation learning methods to capture multimodal demonstrations have been developed in several paradigms. Implicit behavior cloning (IBC) \citep{florence2021implicit} use energy-based models to better capture the action distribution. BeT \citep{shafiullah2022behavior} used the high modeling capabilities of transformers with clustering while DDPM-GPT \citep{pearce2023imitating} leverages transformers alongside discrete-time diffusion models to better fit the action distribution, while BESO \citep{reuss2023goalconditioned} apply discrete-time diffusion while VAE-ACT \citep{zhao2023learning}, DDPM-ACT \citep{chi2024diffusion} added action chunking. While those methods capture diversity, they do so at the a local scale, while we aim to reproduce diversity at the trajectory scale.

\textbf{Offline skill discovery} Offline primitive skill discovery \citep{laskin2022cic, villecroze2022bayesian} is relatively similar to our setting. While its goal is to learn distinct behaviors from a pre-collected dataset of experiences, the result is a set of skills that can be used to improve the agent's efficiency, enable transfer learning, or support hierarchical RL. Such methods aim at making meaningful skills emerge from offline data, but not necessarily capture the whole trajectory distribution.

\textbf{Sequential decision making as an autoregressive generative process} Seeing environment interaction as an autoregressive generation process of trajectories has seen success in the past few years with the growing use of generative models such as transformers \citep{chen2021decision,janner2021sequence} and diffusion models \citep{janner2022planning, ajay2023is} to solve RL tasks. While the main objective of those papers remained the optimization of a reinforcement learning policy, their goal was to step away from Temporal difference (TD) learning and solve RL through reward conditioned trajectory generation, using return-to-go conditioning \citep{chen2021decision}, rewards maximization through beam search \citep{janner2021sequence} or classification-free guidance \citep{janner2022planning,ajay2023is}. Our work shares the same vision but with the objective to regenerate the actual demonstration trajectory distribution.

\textbf{Multimodal imitation learning at the trajectory-scale} Capturing diverse behaviors at trajectory scale has been also studied through the scope of IRL using adversarial frameworks as in \cite{hausman2017multimodal,wang2017robust} and \cite{li2017infogail} while allowing interaction with the environment. In our framework, we aim to capture such diversity in a fully offline manner. In the offline setting, \cite{yang2025diverse} propose to leverage hard coded trajectory labels to learn a finite set of stylized policies, which is different than our approach that aims for learning diverse behaviors at a trajectory scale and in an unsupervised manner. \cite{mao2024stylized} proposed to use Expectation-Maximization (EM) algorithm to build a without supervision a finite set of policies that exhibit diverse behaviors while performing in RL tasks, while we aim to learn a continuum of styles for trajectory diversity capture.

\section{Stylized Imitation Learning for Robust Diverse Controllable Behavior
Capture}
In this part, we:
\begin{enumerate}
    \item Define the notion of trajectory, transition-level and trajectory-level diversity.
    \item Introduce an algorithm to capture the demonstration diversity at trajectory-level based on a latent encoding called style.
    \item Introduce a new framework to relax our method in order to achieve better robustness in stochastic configurations.
\end{enumerate}

\subsection{Transition and trajectory scale diversity} \label{Transition and trajectory scale diversity}
Training agents from demonstrations to achieve certain tasks in an environment has been a key topic in the sequential decision making literature. In this setting, we model the environment as a (rewardless) Markov decision process (MDP) \(\mathcal{M} = (\mathcal{S},\mathcal{A},p_0(s),p(s'|s,a))\) 
defined by a state space \(\mathcal{S}\), an action space \(\mathcal{A}\), an initial state distribution \(p_0(s)\) and a transition kernel $p(s'|s,a)$.  
The agent interacts with the environment according to a policy \(\pi(a|s)\) in an autoregressive sequential generation process of state-action sequences. First, the agent is initialized in a state \(s_0 \sim p_0(\cdot)\) and an action is sampled considering the initial state \(a_0 \sim \pi(\cdot|s_0)\). Then, the environment transitions in a new state \(s_1 \sim p(\cdot|s_0,a_0)\) and the process repeats itself until a final state is reached \(s_T \sim p(\cdot|s_{T-1},a_{T-1})\). As such, we can define \textbf{trajectories} as sequence of states and actions as: \(\tau = (s_0, a_0, s_1, ..., s_{T-1}, a_{T-1}, s_T) \in \mathcal{T}\). A couple \((\mathcal{M},\pi)\) defines consequently a probability distribution on trajectories noted \(p_{\mathcal{M},\pi}(\tau)\). 

In this work, like previous work, we assume that we have access to a dataset \(\mathcal{D}_e\) generated by a set of stylized expert policies \(\Pi_e = \{\pi_e^{(1)}, \pi_e^{(2)}, \ldots, \pi_e^{(K)}\}\) of an unknown number \(1\leq K \leq |\mathcal{D}_e|\) and representing distinct behaviors. Trajectories of \(\mathcal{D}_e\) were generated by first sampling a policy \(\pi_e^{(k)}\) from \(\Pi_e\) according to an unknown distribution \(\mu(\pi)\), and then generating a trajectory \(\tau\) from \(p_{\mathcal{M}, \pi_e^{(k)}}(\tau)\). We note the resulting distribution \(p_{\mathcal{M}, \mu}(\tau)\). Traditional IL methods aim to learn a policy to optimize the likelihood \(\pi(a_t|s_t)\) of \(\mathcal{D}_e\)'s actions given its states, hence focusing on capturing the \textbf{transition-level} diversity of \(\mathcal{D}_e\). While this approach would indeed be sufficient to learn \(\pi\) such that \(p_{\mathcal{M},\pi}(\tau) \approx p_{\mathcal{M},\mu}(\tau)\) in the case of \(k=1\), this would fall short for human generated data which present high level of behavior diversity \((k \gg 1)\). Learning a unique policy would indeed result in averaging the different behavior modes without taking into account \textbf{trajectory-scale} behaviors (e.g. always choosing the same option given a sequence of choices).

\subsection{Style as a trajectory scale conditioning}
As capturing the transition-level multimodality of a dataset is merely a prerequisite for capturing trajectory-level multimodality, many transition-level (BC) \citep{NIPS1988_812b4ba2} fail to imitate multimodal behaviors which are in fact predominant in many settings such as human demonstrations. Hence, it is paramount to give long term temporal information to the policy to condition it to perform diversity capture at trajectory-level. While it is possible to give to the policy a complete or partial history of its behavior, this would lead to an increase of inference time along the trajectory (in the case of transformers), or would lead to catastrophic forgetting in the case of sequential models such as RNNs. Either way, this would also make the generation process of the trajectory uncontrollable. In this work, we propose to condition our policy \(\pi(a|s,z)\) to a latent encoding \(z \in \mathcal{Z}\) of a target trajectory called \textbf{style}. The style aims to represent a latent encoding of the trajectory, projecting all trajectory information in a continuous latent space of low dimension. For this, a classical VAE-like approach would be to train jointly two neural networks: \(e_\phi(z|\tau)\), the trajectory encoder network, and \(\pi_\theta(a|s,z)\), the conditioned policy, to optimize the ELBO of demonstrations, based on trajectory-level styles sampled from the global encoder $e_\phi$. However, achieving accurate encoders of full trajectories as in \cite{wang2017robust} is very challenging, and resources demanding, with prohibitive costs in many high-dimensional applications (e.g., video games from sequences of images).  

Rather, we build on a simpler approach where the encoder is replaced by an embedding matrix that deterministically associates a trainable embedding vector to each trajectory index for a given dataset of indexed trajectories \(\mathcal{D} = \{\tau_i\}\): \(e_\phi(z|\tau_i) = \delta_{z_i}(z)\). In this setting (called ZBC hereafter), we consequently aim to minimize the following loss function:
\begin{align}
    \mathcal{L}_{ZBC}(\phi,\theta) = - \mathbb{E}_{\tau_i \sim \mathcal{D}} \left [ \mathbb{E}_{(s_t^i,a_t^i)\sim \tau_i} \left [ \log \pi_\theta(a_t^i | s_t^i, z_i) \right ] \right ]
\end{align}
With ZBC, we can simply regenerate \(p_{\mathcal{M},\pi}(\tau) \approx p_{\mathcal{M},\mu}(\tau)\) by sampling uniformly a latent vector from the style cookbook \(\{z_i, i=0,...,|\mathcal{D}|-1\}\):
\begin{align}
    p_{\mathcal{M},\pi}^{ZBC}(\tau)  = \frac{1}{|\mathcal{D}|}\sum_{i=0}^{|\mathcal{D}|-1}p_{\mathcal{M},\pi(\cdot|\cdot,z_i)}(\tau)
\end{align}
\subsection{Similarity weighted regression}
While solving tasks within an MDP as been increasingly tackled in the literature \citep{chen2021decision,janner2021sequence} as a sequential generation process, unlike text or videos, it is done through the interaction of a controllable policy and a stochastic and unknown environment. Consequently, offline RL and IL methods might suffer because of two aspects: policy error accumulation as well as environment stochasticity induced by \(p_0(s)\) and \(p(s'|s,a)\). Those can lead our agent to generate a trajectory that drifts away from the initial target trajectory, leading to an unwanted trajectory and possibly out-of-distribution \((s,z)\) configurations where there is no guaranty of the optimality of our actions.

More precisely, this could hurt our model performance in two ways:
\begin{itemize}
    \item Task completion: As ZBC tends to overfit, the non-optimality of the actions could lead to a failure of the task.
    \item Style control: If our policy generalizes well enough on unseen input couple \((s,z)\), it might still perform actions that lead to a very different trajectory from the control.
\end{itemize}

To solve the dichotomy between the robustness of a standard BC and the controllability of ZBC, we propose a novel algorithm that relaxes ZBC in a framework called similarity weighted regression (SWR). We introduce an intermediary method between BC and ZBC called WZBC which allows to capture controllable diversity while being more robust to environment stochasticity. We introduce the notion of trajectory dissimilarity. Given a set of trajectories \(\mathcal{D} = \{\tau\}\), we call dissimilarity a symmetric function \(\nu: \mathcal{T} \times \mathcal{T} \rightarrow [0,1]\) such that:
\begin{align}
    \forall \tau \in \mathcal{D}, \nu(\tau,\tau) = 0 \;\; \mathrm{and} \; \underset{\tau' \in \mathcal{D}\backslash\{\tau\}}{\mathrm{max}} \; \nu(\tau,\tau') = 1
\end{align}
\newpage
This leads us to consider the WZBC approach  given in algorithm~\ref{alg:cap}, which learns the policy and the control space by sampling  couples of trajectories $(\tau_i,\tau_j)$ from the dataset, and weighting the cloning of actions from $\tau_i$ with $\pi_\theta$ conditioned with a style embedding from $\tau_j$,  regarding the dissimilarity $\nu(\tau_i,\tau_j)$. In our experiments we considered the following weighting scheme, with $\beta$ an hyper-parameter tuning the bandwidth of respective influence between trajectories: 
\begin{align}
    \forall (\tau_i,\tau_j), W_{\beta,\nu}(\tau_i,\tau_j) = \exp(- \beta \cdot \nu(\tau_i,\tau_j)), \beta \geq 0
\end{align}
We note that BC can be retrieved by setting \(\beta = 0\) for all dissimilarities \(\nu\) and ZBC by setting \(\nu(\tau_i=\tau_j) = \mathbbm{1}(\tau_i\neq\tau_j)\) and \(\beta \gg 1\). As we seek to discover a set of style embeddings that encode specific behaviors from the dataset, we stop gradients on $e_\phi$ for samples belonging to other trajectories. That way, only the policy is trained globally, leveraging from mixing styles of similar behaviors, while style variables capture their specific control semantics. 
\begin{algorithm}
\caption{WZBC}
\label{alg:cap}
\begin{algorithmic}
\State \textbf{Input:} Offline dataset of trajectories $\mathcal{D}$
\State \textbf{Precompute similarities (for faster training):}
\For{\((\tau_i,\tau_j) \in \mathcal{D}\)}
    \State Compute and store \(\nu_{i,j} = \nu(\tau_i,\tau_j)\) 
\EndFor
\textbf{Train policy:}
\While{not converged}
    \State Sample a couple of trajectories \((\tau_i,\tau_j)\) of trajectories from \(\mathcal{D}\) such that \(i \neq j\) with probability \(p\)
    \State Sample a transition \((s_t^i,a_t^i)\) from \(\tau_i\) and a style \(z_j\) from \(e_\phi(z|\tau_j)\)
    \If{\(i = j\)}
        \State  \((\theta,\phi) \leftarrow (\theta,\phi) + \lambda \nabla (W_{\beta,\nu}(\tau_i,\tau_j) \log \pi_\theta(a_t^i|s_t^i,z_j))\)
    \Else
        \State  \((\theta,\phi) \leftarrow (\theta,\phi) + \lambda \nabla (W_{\beta,\nu}(\tau_i,\tau_j) \log \pi_\theta(a_t^i|s_t^i,\textbf{sg}[z_j]))\) (We stop gradients of relabeled styles)
    \EndIf
\EndWhile
\end{algorithmic}
\end{algorithm}
\\
The choice of the dissimilarity metric 
depends on the use cases. 
In the following, we chose 
to consider a simple euclidean distance between trajectory states of same timestep through padding the trajectories to the same length by repeating the last state in the sequence to make them comparable: 
\begin{align}
    \forall (\tau_1,\tau_2) \in \mathcal{D}, \nu(\tau_1,\tau_2) = \frac{||\textbf{pad}(\tau_1^s) - \textbf{pad}(\tau_2^s)||}{\underset{\tau_3 \in \mathcal{D}}{\max}||\textbf{pad}(\tau_1^s) - \textbf{pad}(\tau_3^s)||}
\end{align}


\section{Experiments}
Our experiments will be carried out on 4 tasks to answer the following questions: 
\begin{enumerate}
    \item How does ZBC and WZBC perform in diversity reconstruction compared to prior works ?
    \item How much control on the generated trajectories can we get from our methods ?
    \item How robust are methods in the case of stochastic environments ?
\end{enumerate}

\subsection{Experimental setup}
To illustrate the benefits of our approach, we carry out our experiment on a set of diverse human generated datasets in several environments. 
\textbf{Maze2D} is our set of 2d mazes from which we generated navigation datasets with various paths. \textbf{D3IL} \citep{jia2024diverse} is a suite of robotic tasks with human generated diverse datasets. Each environment has a finite number of tractable behaviors \(\mathcal{B}\).
\subsubsection{Maze2D}
\begin{figure*}[h!]
    \centering
    \begin{subfigure}[b]{0.23\textwidth}
        \centering
        \includegraphics[width=\textwidth]{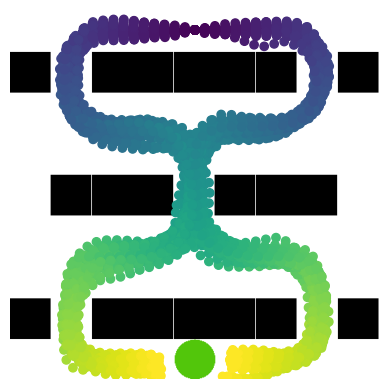}
        \caption{One side}
    \end{subfigure}
    \hfill
    \begin{subfigure}[b]{0.23\textwidth}
        \centering
        \includegraphics[width=\textwidth]{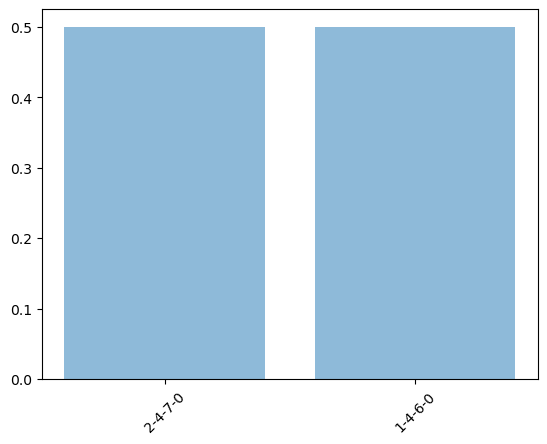}
        \caption{One side histogram}
    \end{subfigure}
    \hfill
    \begin{subfigure}[b]{0.23\textwidth}
        \centering
        \includegraphics[width=\textwidth]{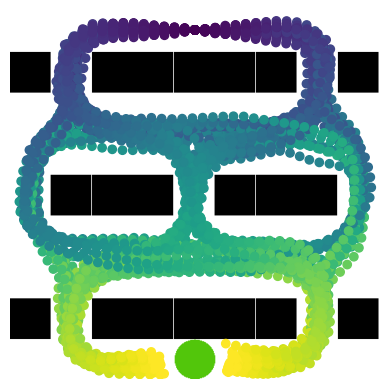}
        \caption{Only forward}
    \end{subfigure}
    \hfill
    \begin{subfigure}[b]{0.23\textwidth}
        \centering
        \includegraphics[width=\textwidth]{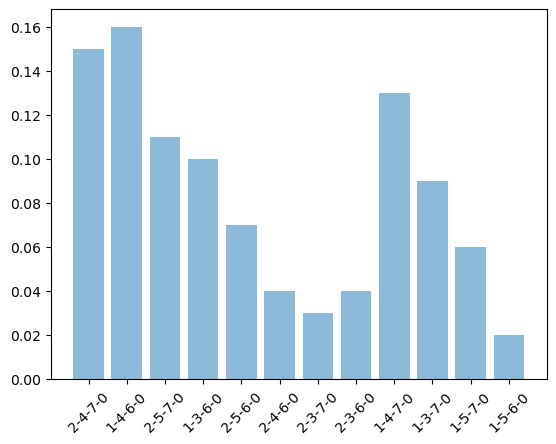}
        \caption{Only forward histogram}
    \end{subfigure}

    \caption{(Left) Trajectories in the maze: the start of a trajectory is shown in blue, 
      the end in yellow, and the goal in green. (Right) Histograms of the 
      behavior distribution of the datasets.}
    \label{fig:maze_datasets}
\end{figure*}
The \textbf{Maze2D} environment suite is a set of simple and fast 2D navigation environments where the goal is to go from a starting point to a goal point by navigating through several doors in the  mazes with different sizes and dynamics (noisy transitions, sticky walls, various initialization states distribution). The behaviors corresponds to the sequences of checkpoints visited in the trajectory. Our experiments focused on the \textbf{medium\_maze} environment. Each door represent an indexed checkpoint which is activated once it is visited, while the goal itself represents the checkpoint \(0\). A trajectory corresponds to a sequence of checkpoints, leading to a high number of behaviors \(|\mathcal{B}| \in \mathbb{N}\). We provide 2 different types of datasets, displayed in Fig. \ref{fig:maze_datasets}: 
\paragraph{One side} The One side dataset contains two types of trajectories: right and left trajectories, joining in the center. Hence, for medium\_maze: \(\mathcal{B} = \{6410,7420\}\) and \(k=|\mathcal{B}|=2\). This dataset aims at checking the capacity of algorithms to capture the temporal consistency of the diversity of a dataset at trajectory-level. It contains 100 trajectories, with the same amount of trajectories for each behavior.
\paragraph{Only forward} The Only Forward dataset aims at evaluating the capacity of our algorithms to capture diverse means of navigating the map. It displays \(k=|\mathcal{B}| = 12\) behaviors within 100 trajectories, with an unbalanced number of trajectory for each behaviors.
\subsubsection{D3IL}
\begin{figure}[h!]
    \centering
    \begin{subfigure}[t]{0.22\textwidth}
        \centering
        \includegraphics[width=\textwidth]{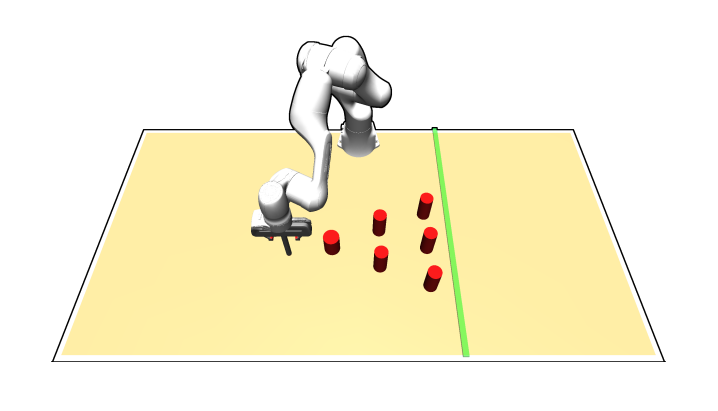}
        \caption{Avoiding}
    \end{subfigure}
    \hfill
    \begin{subfigure}[t]{0.22\textwidth}
        \centering
        \includegraphics[width=\textwidth]{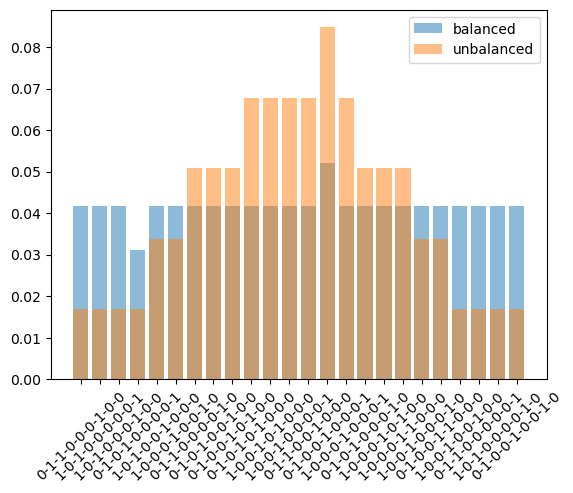}
        \caption{Avoiding histogram}
    \end{subfigure}
    \hfill
    \begin{subfigure}[t]{0.22\textwidth}
        \centering
        \includegraphics[width=\textwidth]{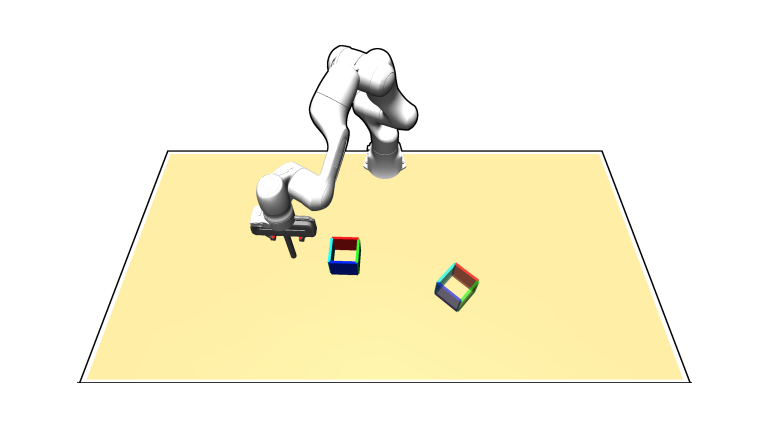}
        \caption{Aligning}
    \end{subfigure}
    \hfill
    \begin{subfigure}[t]{0.22\textwidth}
        \centering
        \includegraphics[width=\textwidth]{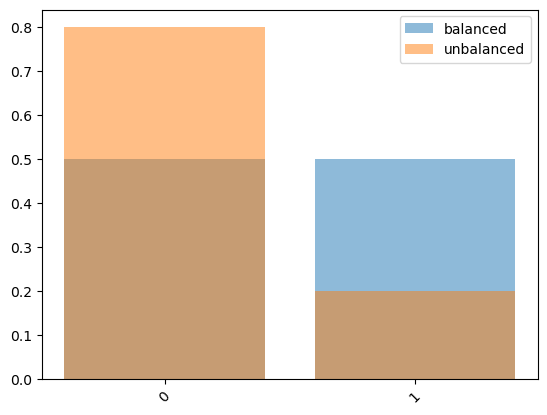}
        \caption{Aligning histogram}
    \end{subfigure}
    \hfill
    \caption{(Left) Pictures of the environments. (Right) Histograms of 
      the behavior distribution of the datasets. In blue are the 
      provided dataset’s behaviors and in yellow are those 
      of our unbalanced dataset.}
    \label{d3il_datasets}
\end{figure}
The D3IL environments are illustrated in Fig. \ref{d3il_datasets} and correspond the the following:
\paragraph{Avoiding} In this task, the robot must travel from a fixed starting position to a green finish line without colliding with any of six obstacles. Since no object manipulation is involved, the primary challenge is capturing a wide range of behaviors. There are 24 distinct successful paths (\(k=|\mathcal{B}| = 24\)). The dataset contains 96 demonstrations, comprising four trajectories for each of the 24 solutions.
\paragraph{Aligning} Here, the robot is required to push a hollow box to a specified position and orientation. This can be done from inside or outside the box, resulting in two possible behaviors (\(k=|\mathcal{B}| = 2\)). Although it demands less behavioral diversity than the Avoiding task, Aligning involves more complex object manipulation. The dataset includes 1,000 demonstrations, 500 for each behavior, collected over uniformly sampled initial states.

Because these datasets were curated to exhibit uniformly distributed behaviors, we additionally generated unbalanced versions to test how well our algorithms can reproduce particular behavior distributions. Fig.~\ref{d3il_datasets} illustrates the resulting behavior histograms.

\subsection{Visualizing sampling distributions}
\begin{figure}[h!]
    \centering
    \begin{subfigure}[t]{0.3\textwidth} 
        \centering
        \includegraphics[width=\textwidth]{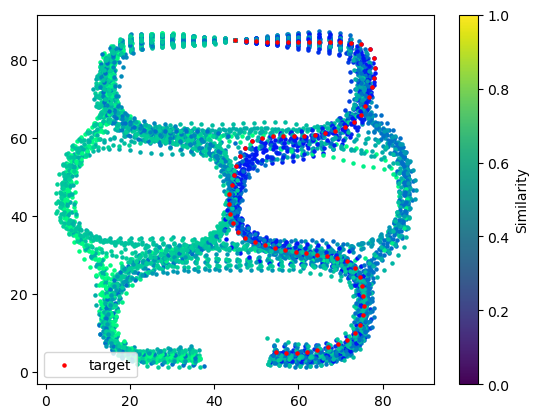}
        \caption{Case 1}
    \end{subfigure}
    \hfill
    \begin{subfigure}[t]{0.3\textwidth}
        \centering
        \includegraphics[width=\textwidth]{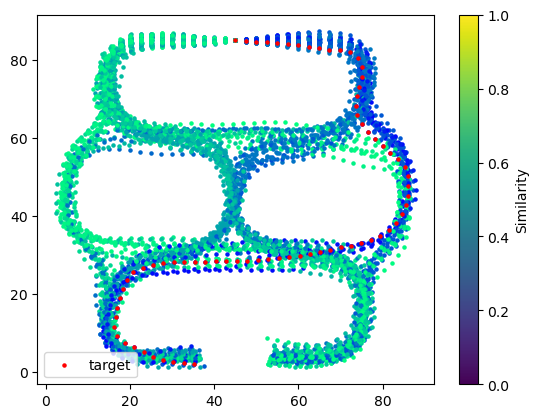}
        \caption{Case 2}
    \end{subfigure}
    \hfill
    \begin{subfigure}[t]{0.3\textwidth}
        \centering
        \includegraphics[width=\textwidth]{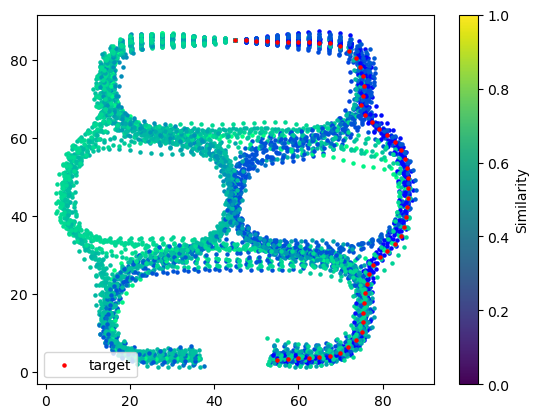}
        \caption{Case 3}
    \end{subfigure}

    \caption{\textbf{Dissimilarity values of trajectories} 
      \(\nu({\color{red} \tau^*},\tau)\) for different reference 
      trajectories \({\color{red} \tau^*}\) in red. 
      Blue trajectories are the most similar, 
      green the most dissimilar.}
    \label{fig:horizontal_plots}
\end{figure}
\begin{figure}[h!]
    \centering
    \begin{subfigure}[t]{0.3\textwidth}
        \centering
        \includegraphics[width=\textwidth]{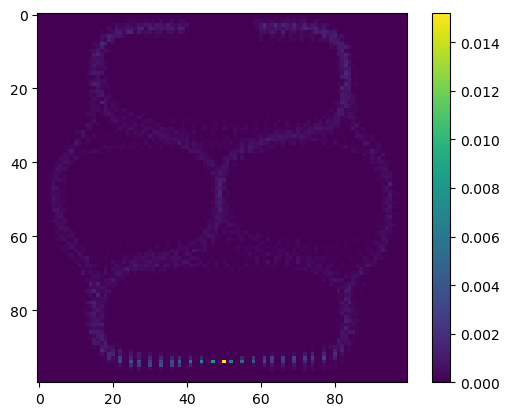}
        \caption*{\(\beta = 0\)}
    \end{subfigure}
    \hfill
    \begin{subfigure}[t]{0.3\textwidth}
        \centering
        \includegraphics[width=\textwidth]{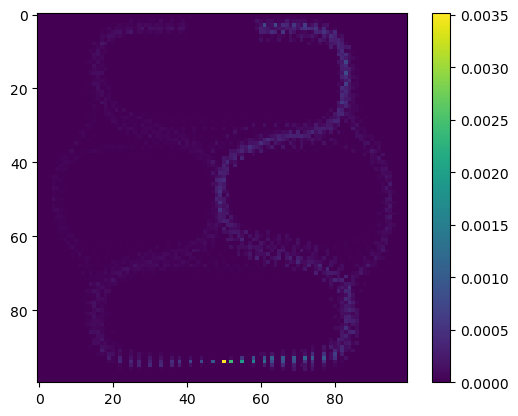}
        \caption*{\(\beta = 3.0\)}
    \end{subfigure}
    \hfill
    \begin{subfigure}[t]{0.3\textwidth}
        \centering
        \includegraphics[width=\textwidth]{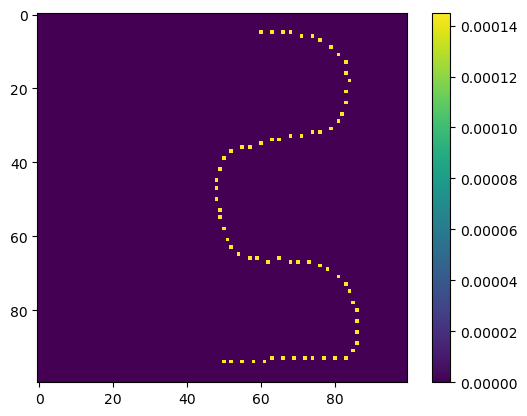}
        \caption*{\(\beta = 100.0\)}
    \end{subfigure}
   \caption{\textbf{Values of the conditional input sample distributions \(\rho(\tilde{s}|z)\)}. \(\beta = 0\) give similar input sample distribution as BC, while \(\beta = 100\) give a similar input sampling distribution as ZBC. We can find a middle ground with \(\beta = 3.0\), allowing all the support of BC but with a significant weighting to distinguish the trajectories similarities.}
  \label{values_conditional}
\end{figure}
To visualize the impact of the SWR, we compute the similarity weighted probability of each states sample during training defined as \(\rho_\mathcal{D}(s|z)\) for all \(s \in \mathcal{S}\) given a certain \(z \in \mathcal{Z}\). Several methods for density estimation are available to us. Traditionally, Parzen windows were used for fast log-likelihood estimation, but their estimation quality have been shown to be unreliable \citep{theis2016note}. Normalizing flow methods are also available but they can be expensive to train. We propose a grid-based approach for density estimation, which finds itself sufficient for visualization purposes. For this, we discretize uniformly the observation space per dimension, giving us a set of categories \(\Tilde{S} = \left \{\text{cat}(s), s \in S \right \}\).
We have consequently:
\begin{equation} \label{eq3}
\begin{split}
\rho_\mathcal{D}(\Tilde{s}|z_i) & = \sum_{j=0}^{|\mathcal{D}|-1}\sum_{t=0}^{|\tau_j|-1}\frac{\mathbbm{1}\{s_t^{\tau_j} \in \tilde{s}\}}{|\mathcal{D}||\tau_j|}\times W(\tau_i,\tau_j)
\end{split}
\end{equation}
We show in Fig.~\ref{fig:horizontal_plots} the values of the dissimilarity of trajectories for 3 different reference trajectories in the Maze2D environment.
We see that for each trajectory, the dissimilarity is well captured along the trajectories, allowing to distinguish between actions of similar or dissimilar trajectories at a given state. We also visualize in Fig.~\ref{values_conditional} the distributions \(\rho(\Tilde{s}|z)\) for a fixed value of \(z \in \mathcal{Z}\) and different values of \(\beta\).
\subsection{Assessing the diversity at trajectory-scale}
\begin{table}
  \centering
  \small
  \renewcommand{\arraystretch}{1.1} 
  \caption{L1 Distance of histograms of sampled behavior with respect to the reference histogram and success rates.}
  \label{tab:l1_distance} 
  \resizebox{\columnwidth}{!}{%
  \begin{tabular}{|l|c|c|c||c|c|c|}
  \hline
  \textbf{Dataset (distance)} & \textbf{BC} & \textbf{ZBC} & \textbf{WZBC} & \textbf{BESO} & \textbf{DDPM-ACT} & \textbf{DDPM-GPT} \\ 
  \hline \hline
  medium\_maze-only\_forward  & 1.74 ± 0.054 & \textbf{0.256 ± 0.023} & \textbf{0.248 ± 0.047} & 0.744 ± 0.041 & 0.916 ± 0.252 & 0.604 ± 0.082 \\ 
  medium\_maze-one\_side      & 1.4 ± 0.49 & \textbf{0.044 ± 0.032} & \textbf{0.06 ± 0.033} & 0.140 ± 0.049 & 0.640 ± 0.390 & 0.100 ± 0.075 \\ 
  \hline \hline
  d3il\_avoiding             & 1.917 ± 0.0 & \textbf{0.265 ± 0.0} & 0.482 ± 0.026 & 0.901 ± 0.091 & 0.781 ± 0.184 & 0.531 ± 0.093 \\ 
  d3il\_unbalanced\_avoiding & 1.925 ± 0.062 & \textbf{1.02 ± 0.116} & 1.457 ± 0.087 & 1.283 ± 0.067 & 1.342 ± 0.134 & 1.26 ± 0.026 \\ 
  d3il\_aligning             & 1.0 ± 0.0 & \textbf{0.172 ± 0.17} & 0.552 ± 0.224 & 0.472 ± 0.111 & 0.488 ± 0.075 & 0.296 ± 0.104 \\ 
  d3il\_unbalanced\_aligning & 0.4 ± 0.0 & \textbf{0.172 ± 0.057} & 0.364 ± 0.037 & 0.256 ± 0.066 & 0.212 ± 0.053 & 0.288 ± 0.063 \\
  \hline \hline
  \textbf{Dataset (success rate)} & \textbf{BC} & \textbf{ZBC} & \textbf{WZBC} & \textbf{BESO} & \textbf{DDPM-ACT} & \textbf{DDPM-GPT} \\ 
  \hline \hline
  medium\_maze-only\_forward  & 1.0 ± 0.0 & 1.0 ± 0.0 & 0.99 ± 0.0 & 0.998 ± 0.004 & 0.9 ± 0.12 & 1.0 ± 0.0 \\ 
  medium\_maze-one\_side      & 0.6 ± 0.49 & 1.0 ± 0.0 & 1.0 ± 0.0 & 1.0 ± 0.0 & 0.994 ± 0.012 & 1.0 ± 0.0 \\ 
  \hline \hline
  d3il\_avoiding             & 1.0 ± 0.0 & 0.996 ± 0.005 & 0.954 ± 0.024 & 0.998 ± 0.004 & 0.994 ± 0.008 & 0.986 ± 0.006 \\ 
  d3il\_unbalanced\_avoiding & 0.6 ± 0.49 & 0.75 ± 0.092 & 0.802 ± 0.113 & 1.0 ± 0.0 & 0.99 ± 0.013 & 0.996 ± 0.005 \\ 
  d3il\_aligning             & 0.21 ± 0.395 & 0.552 ± 0.032 & 0.806 ± 0.105 & 0.908 ± 0.012 & 0.872 ± 0.047 & 0.852 ± 0.055 \\ 
  d3il\_unbalanced\_aligning & 1.0 ± 0.0 & 0.328 ± 0.054 & 0.762 ± 0.126 & 0.922 ± 0.013 & 0.882 ± 0.038 & 0.844 ± 0.015 \\ 
  \hline
  \end{tabular}
  }
\end{table} 
To measure the fidelity of our reconstructions, we need to define a way to quantify the similarity between \(p_{\mathcal{M},\pi}(\tau)\) and \(p_{\mathcal{M},\mu}(\tau)\).
Many methods quantified diversity of the generated trajectories through the entropy of some behavior histograms \citep{mao2024stylized,jia2024diverse}. It can be shown \citep{shannon} that entropy is maximal for the uniform distribution. As such, those methods were evaluated on how they maximize diversity in an uniform way, and not reproduce the diversity of human demonstrations. It is notable that the recent study in \citep{jia2024diverse} tackled this issue by focusing on uniformly distributed human data, but with the loss of generality on their evaluation, human data being most of the time biased towards certain displays of behaviors. Rather, to quantify the diversity of the trajectory distributions, we propose to analyze a discretized distribution \(h\) (or histogram) of some defined \textbf{metrics} \(m_b\) of the trajectories (length, sequence of checkpoints, cumulated rewards, ...). Mazes are assessed through sequences of checkpoints, D3IL environments consider specific behavior metrics for each task. To quantify the distance between  \(p_{\mathcal{M},\pi}(\tau)\) and \( p_{\mathcal{M},\mu}(\tau)\), we use the L1-distance between these corresponding behavior histograms with zero-padding to match the supports of the histograms. We compare our methods with algorithms from the D3IL \citep{jia2024diverse} benchmark: BESO \citep{reuss2023goalconditioned}, DDPM-ACT \citep{chi2024diffusion} and DDPM-GPT \citep{pearce2023imitating}. In Table \ref{tab:l1_distance}, we see that ZBC is performing the best in behavior diversity reconstruction, which highlights the benefits from global control that is simply introduced by our base method,  while suffering from a lack of performance in the D3IL tasks. WZBC is competitive in behavior L1 diversity reconstruction for medium\_maze datasets to ZBC but not in the D3IL environments. However, WZBC performs in D3IL better than ZBC (see success rate for tasks in Table \ref{tab:l1_distance}). 
\subsection{Control}
\begin{figure}[H]
    \centering
    
    \begin{subfigure}{0.45\textwidth}
        \centering
        \includegraphics[width=\textwidth]{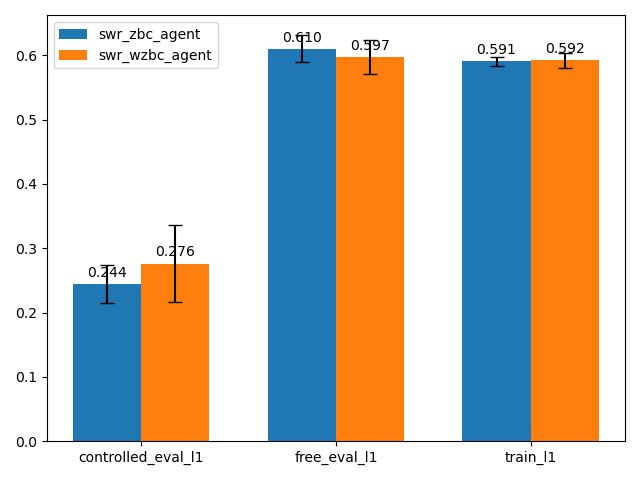}
    \end{subfigure}
    \hfill
    \begin{subfigure}{0.5\textwidth}
        \centering
        \includegraphics[width=\textwidth]{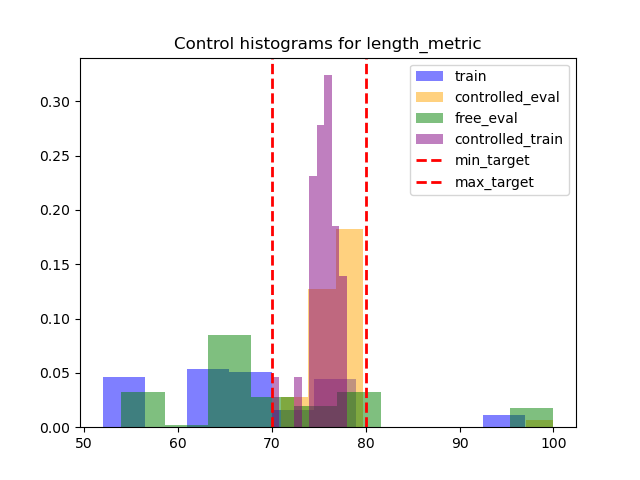}
    \end{subfigure}

    \caption{(Left) L1 distance between the training behavior histogram and respectively: the property controlled agent evaluation behavior histogram, the free agent evaluation  histogram (without filtering controls for desired lengths)  and the controlled train histogram. (Right) In blue: the training length histogram, purple: the conditioned training length histogram, green: the free agent eval length histogram, yellow: the controlled eval length histogram.}
  \label{control_histograms}
\end{figure}
One benefit of style based methods is their controllability. We aim to condition the generation by certain properties on trajectories. Many previous diverse imitation learning methods focused on capturing diversity without considering controllability as a key feature of the learned model. Performing trajectory generation conditioned by a general unsupervised latent encoding allows to condition the generation to various criteria, such as defined metric values, without needed any finetuning in low stochasticity scenarios. In our models, it is possible to condition the autoregressive generation on every given trajectory property \(\Psi(\tau) \in \{\text{True},\text{False}\}\) , one can generate trajectories according to:
\begin{align}
    p_{\mathcal{M},\pi}(\tau|\Psi(\tau)\text{ is True}) = \int_z p_{\mathcal{M},\pi}(\tau|z)e(z|\Psi(\tau)\text{ is True})dz.
\end{align}
where \(e\) the style sampling distribution. In this paper, we consider controllability conditioned by properties of the from: \(\Psi(\tau) = m(\tau) \in [m_{min},m_{max}]\). We estimate \(e(z|\Psi(\tau))\) by the following:
\begin{align}
    e(z|\Psi(\tau)\text{ is True }) = \frac{\sum_{\tau \in \mathcal{D}} e(z|\tau) \mathbbm{1}(\Psi(\tau)\text{ is True})}{\sum_{\tau \in \mathcal{D}} \mathbbm{1}(\Psi(\tau)\text{ is True})}
\end{align}
Depending on the encoder distribution, this might result in a mixture of dirac distributions or a mixture of gaussians in the latent space. Methods such as Kernel Density Estimation (KDE) could lower the number of gaussian distributions in the mixture, but at a price of approximation fidelity. Such methods could be required nonetheless as the dataset grows in trajectories. In \textbf{medium\_maze-only\_fast-human}, we can condition our trajectories to respect a certain metric as \(\Psi(\tau) = \text{length}(\tau) \in [70,80]\). In practice, we set:
\begin{align}
    e(z|\Psi(\tau)\text{ is True}) = \frac{\sum_{i=0}^{|\mathcal{D}|-1} \delta_{z_i}(z) \mathbbm{1}(\Psi(\tau_i)\text{ is True})}{\sum_{i=0}^{|\mathcal{D}|-1} \mathbbm{1}(\Psi(\tau_i)\text{ is True})} 
\end{align}
We compare the behaviors generated by the controlled policy, by the free policy and that of the whole dataset to the behaviors of the property restricted dataset (all trajectories such that \(\Psi(\tau) = \text{length}(\tau) \in [70,80]\) is true). In Fig.\ref{control_histograms} are displayed the L1 distance as well as the behavior histograms free of conditioning and under conditioning. We see that without further training, we can condition our policy to regenerate trajectories according to the training distribution, conditioned to certain metric properties.
\subsection{Robustness}
\begin{table*}[h!]
\caption{L1 Distance and Success Rate Comparison}
\label{distance_success_robust}
\begin{center}
\renewcommand{\arraystretch}{1.2}  
\setlength{\tabcolsep}{6pt}  
\resizebox{\textwidth}{!}{%
\begin{tabular}{|l|c|c||c|c|}
\hline
\multirow{2}{*}{\textbf{Configuration}} & \multicolumn{2}{c||}{\textbf{L1 Distance}} & \multicolumn{2}{c|}{\textbf{Success Rate}} \\ 
\cline{2-5}
 & \textbf{ZBC} & \textbf{WZBC} & \textbf{ZBC} & \textbf{WZBC} \\ 
\hline
medium\_maze-only-forward (determinist) & $0.256 \pm 0.023$ & $0.248 \pm 0.047$ & $1.0 \pm 0.0$ & $0.99 \pm 0.0$ \\ 
medium\_maze-only-forward (pseudo-r-init) & $1.152 \pm 0.094$ & $0.828 \pm 0.349$ & $0.448 \pm 0.031$ & $0.684 \pm 0.152$ \\ 
medium\_maze-only-forward (r-init) & $1.556 \pm 0.079$ & $1.552 \pm 0.037$ & $0.858 \pm 0.046$ & $0.978 \pm 0.019$ \\ 
medium\_maze-only-forward (noise-transi) & $0.729 \pm 0.134$ & $0.744 \pm 0.029$ & $0.632 \pm 0.066$ & $0.744 \pm 0.038$ \\ 
\hline
\end{tabular}%
}
\end{center}
\end{table*}
We are measuring the performances of ZBC and WZBC in various stochastic contexts: random initialization around the usual starting point (pseudo-r-init), fully random initialization (r-init) and noisy transitions (noise-transi) on medium\_maze-only-forward. As displayed on Table \ref{distance_success_robust}, we see that WZBC outperforms ZBC in most stochastic configurations of the maze environment, improving thus its robustness to stochasticity and temporal distribution shifts, which emphasizes the benefits from leveraging mixes of styles during training.

\newpage
\section{Conclusion}
We introduced a new framework called similarity weighted regression, instantiated by two algorithms: ZBC and WZBC. Those two methods displayed the best performance in diversity capture compared to previous multi-modal imitation learning methods. Those methods were evaluated with the goal to capture the real demonstration diversity at trajectory scale, and not just maximizing the entropy of behaviors, as it is the case in many recent studies. Furthermore, ZBC and WZBC demonstrated good controllability when conditioned by certain metrics of the desired trajectory while being trained in a fully unsupervised manner. Also, on our provided environment, WZBC displayed some robustness to environment stochasticity compared to the baselines.

For future work, as the instantiation of our similarity weighted regression, WZBC, is using an euclidean distance, it might struggle in some high dimensional data scenarios. Consequently, while it corresponds to a good metric for navigating in small mazes and solving various robotics tasks, analyzing more complex similarity metrics for instance based on temporal losses from reinforcement learning value functions that can scale in high dimensional data is a promising research direction.

\section{Implementation details}
The hyperparameter settings for each method are as follows:
\begin{itemize}
    \item \textbf{BC}: hidden\_dim = 128, num\_hidden = 10, batch\_size = 16, lr = $10^{-3}$.
    \item \textbf{ZBC}: hidden\_dim = 128, num\_hidden = 10, batch\_size = 16, lr = $10^{-3}$, style\_dim = 10.
    \item \textbf{WZBC}: hidden\_dim = 128, num\_hidden = 10, batch\_size = 16, lr = $10^{-3}$, style\_dim = 10, $\beta$ = 10.0, $p$ = 0.8.
\end{itemize}
Each algorithm ran on 1e5 gradient steps. The baseline hyperparameters are provided by the D3IL framework. The experiments were conducted on 5 independent seeds.

\section{Acknowledgements}
This work was granted access to the HPC resources of IDRIS under the allocation 2023-AD011014679 made by GENCI.

\bibliography{iclr2024_conference}
\bibliographystyle{iclr2024_conference}

\end{document}